\documentclass[10pt,twocolumn,letterpaper]{article}

\usepackage{cvpr}
\usepackage{times}
\usepackage{epsfig}
\usepackage{graphicx}
\usepackage{amsmath}
\usepackage{amssymb}
\usepackage{amsthm}

\usepackage[breaklinks=true,bookmarks=false]{hyperref}

\cvprfinalcopy 

\usepackage{subfigure}
\usepackage{multirow}
\usepackage{url}
\usepackage{footnote}
\usepackage{array}

\newcommand{\namefull}{Gated Path Selection Network}
\newcommand{\name}{GPSNet}

\def\cvprPaperID{6707} 

\begin{document}

\title{Gated Path Selection Network for Semantic Segmentation}

\author{
Qichuan Geng$^1$\hspace{0.6cm}Hong Zhang\hspace{0.6cm}Xiaojuan Qi$^2$\hspace{0.6cm}Ruigang Yang$^{3}$\hspace{0.6cm}Zhong Zhou$^1$\hspace{0.6cm}Gao Huang$^{4}$
\vspace{0.3cm}\\  
$^1$Beihang University \quad $^2$University of Oxford \quad $^3$University of Kentucky\quad $^4$ Tsinghua University 
\\
{\tt\small \{zhaokefirst,zz\}@buaa.edu.com} \quad 
{\tt\small \{fykalviny,qxj0125\}@gmail.com} \\
\vspace{-0.02in}
{\tt\small Ryang2@uky.edu \quad gaohuang@tsinghua.edu.cn}
}

\maketitle

\begin{abstract}
    Semantic segmentation is a challenging task that needs to handle large scale variations, deformations and different viewpoints. In this paper, we develop a novel network named \namefull{}~(\name{}), which aims to learn adaptive receptive fields. In \name{}, we first design a two-dimensional multi-scale network -- SuperNet, which densely incorporates features from growing receptive fields. To dynamically select desirable semantic context, a gate prediction module is further introduced.
    In contrast to previous works that focus on optimizing sample positions on the regular grids, \name{} can adaptively capture free form dense semantic contexts.
    The derived adaptive receptive fields are data-dependent, and are flexible that can model different object geometric transformations.
    On two representative semantic segmentation datasets, \ie, Cityscapes, and ADE20K, we show that the proposed approach consistently outperforms previous methods and achieves competitive performance without bells and whistles.
\end{abstract}
\section{Introduction}
\label{intro}
Semantic segmentation refers to the problem of assigning semantic object category for each pixel. Recent state-of-the-art semantic segmentation approaches~\cite{fcn,chen2014semantic,chen2017deeplab,chen2017rethinking,chen2018encoder,pspnet,denseaspp} are typically based on the Fully Convolutional Networks~(FCNs)~\cite{fcn}. It benefits from the informative representations of object categories and semantic information learned by Convolutional Neural Networks~(CNNs)~\cite{resnet,densenet}. CNNs are inherently limited by the design of structure, where the receptive field is restricted to constant regions~\cite{psanet}. Objects in semantic segmentation are in a large range of scales, deformations and different viewpoints, and the fixed field of view in CNNs is insufficient to deal with geometric variations. 

Extensive efforts have been made to enlarge receptive fields to better understand the semantic scenes.
Atrous convolution~\cite{chen2014semantic} incorporates larger context by dilating the convolution kernel in a fixed manner. However, it lacks the ability to cope with multi-scale objects. To mitigate the problem, PSPNet~\cite{pspnet} applies~\emph{pyramid pooling module} to aggregate information from different scales of feature maps. ASPP~\cite{chen2017deeplab} and DenseASPP~\cite{denseaspp} are introduced to use a series of atrous convolution layers to learn features with multiple dilation rates. Nevertheless, those approaches ignore the same problem: different regions may correspond to objects with different scales and geometric variations, Therefore, receptive fields on regular grids are insufficient to tackle large appearance variations.

To capture rich semantic context, 
attention-based approaches~\cite{psanet,encnet,ocnet,danet} are proposed to adaptively aggregate short- and long-range features. To spotlight locally discriminative information, the recent work Deformable Convolutional Network~(DCN)~\cite{dcn} shows that the adaptive sample positions can be acquired by predicting additional offsets.
However, the learned receptive field is limited by the size of pre-defined convolution kernel which is difficult to capture long-range context. Moreover, DCN sparsely samples a fixed number of locations rather than considering all relevant locations within the contextual scope.

In this work, we introduce \emph{\namefull{}~(\name{})} to enhance the capability of CNNs to adaptively learn free form receptive fields for semantic segmentation. Different from previous approaches which focus on optimizing sample positions on the regular grids, \name{} densely captures semantic contexts which are adaptively weighted. The learned adaptive receptive fields are data-dependent, and are flexible to model various geometric deformations. Figure~\ref{fig:overview} gives an overview of the proposed method.

Specifically, to construct a set of message passing paths with various receptive fields and sample rates~\cite{denseaspp}, we carefully design a two-dimentional multi-scale network architecture with multiple branches.
We refer to the network as \emph{SuperNet}. It stacks a series of bottlenecked branches which consist of different tuned dilation convolutions. To progressively produce multi-scale features, inspired by \cite{huang2019convolutional, denseaspp}, dense connections are applied to feed the input from relative small receptive fields to the large receptive fields. Moreover, in order to aggregate information from different paths, a gate prediction module is introduced to predict soft masks to combine the semantic contexts. The gates not only enable dynamic selection of the effective pixel positions but also help regularize the receptive fields. Hence, different scales of objects can obtain desirable representations. Moreover, another notable advantage of \name{} is its improved sample rates, which ensures that both the interior and exterior regions can be densely sampled to capture adequate semantic context. 

The GPS module is light-weight as it adds a small amount of parameters and computation for multi-scale feature extraction. 
Meanwhile, it is model-agnostic which can be readily used in various ASPP-like structure and trained in an end-to-end manner.

We conduct experiments on two competitive semantic segmentation datasets, \ie, Cityscapes~\cite{cityscapes}, and ADE20K~\cite{ade20k}. The experiments demonstrate that our proposed \emph{\name{}} consistently improves the performance of previous state-of-the-art approaches.
\section{Related Work}
\label{related}
\paragraph{Semantic segmentation.}
Semantic segmentation is a fundamental problem in computer vision, which involves assigning a semantic category to each pixel. 
Recent progress in this problem has been largely driven by deep fully convolutional neural networks (FCNs)~\cite{fcn,zheng2015conditional,yu2015multi,chen2017deeplab,pspnet}.
The pioneering work FCN ~\cite{fcn} proposed to remove the fully connected layers in classification CNN networks~\cite{KrizhevskySH12,simonyan2014very}, leading to a fully convolutional architecture for dense semantic segmentation. 
To facilitate dense prediction, the deconvolutional layer was also introduced in~\cite{fcn,noh2015learning}, which is a learnable upsampling operation.
The following work SegNet~\cite{badrinarayanan2017segnet} introduced an encoder and decoder network, where the decoder utilizes pooling indices in the encoding layers to upsample the feature map. UNet~\cite{ronneberger2015u} adopted skip connections to combine shallow representations from the encoder and deep features from the encoder, which exploit low level feature for accurate semantic segmentation.
To refine segment contours, CRF was applied as a post-processing procedure~\cite{chen2017deeplab} or end-to-end integrated~\cite{zheng2015conditional} into the network.
Further, to develop real-time semantic segmentation networks for practical applications, Paszke \etal proposed a light weight neural network -- ENet~\cite{paszke2016enet} by exploiting separable convolution and less channels. Zhao \etal~\cite{zhao2018icnet} proposed ICNet to utilize image pyramid to optimize the network.

\paragraph{Contextual information modeling in semantic segmentation.}
Recent work has shown that contextual information is important for improving semantic segmentation accuracy.
Deeplab~\cite{chen2017deeplab} and Dilated Conv~\cite{yu2015multi} proposed atrous convolutions to enlarge the network receptive field without sacrificing the resolution, which enables the network to harvest contextual information in a larger region for semantic segmentation.
Moreover, an atrous spatial pyramid pooling (ASPP) module was developed to incorporate contextual information from multiple scales.
ParseNet~\cite{liu2015parsenet} proposed global average pooling layer which introduces global contextual information for semantic segmentation. 
Later on, Zhao \etal~\cite{pspnet} proposed a Pyramid Pooling Module to aggregate contextual information from multi-scale regions. 
However, the above approaches can only aggregate contextual information from predefined or fixed regions, which still be limited for modeling large contextual variations in objects.

\paragraph{Deformable convolution or attention mechanism in semantic segmentation.}
Deformable and attention based mechanisms enable the network to dynamically select of the context for each pixel, which is adaptive to different pixels.
Chen \etal~\cite{chen2016attention} proposed to learn combination weights for combining multi-scale features. 
Deformable convolution layer is introduced in~\cite{zhu2019deformable}, which makes the convolution kernel adaptive to geometric variations of the object, extracting dynamic contextual information for image recognition.
Wang \etal~\cite{wang2018non} proposed non-local operation that computes a weighted sum of features in the global map based on a attention mechanism. 
Further, Point-wise Spatial Attention Network~\cite{psanet} learns to aggregate information through a learned attention map dynamically adjusting contextual information aggregation. 
OCNet~\cite{ocnet} proposed an object context network to learn an object context map by modeling pixel-pixel similarities which are further utilized to refine the representations of each pixel.
CCNet~\cite{huang2019ccnet} harvested the contextual information on the criss-cross path which can provide long-range contextual information to each pixel with improved efficiency.
\section{\namefull{}}
\label{method}
\begin{figure*}[bpt]
    \centering
    \includegraphics[width=\linewidth]{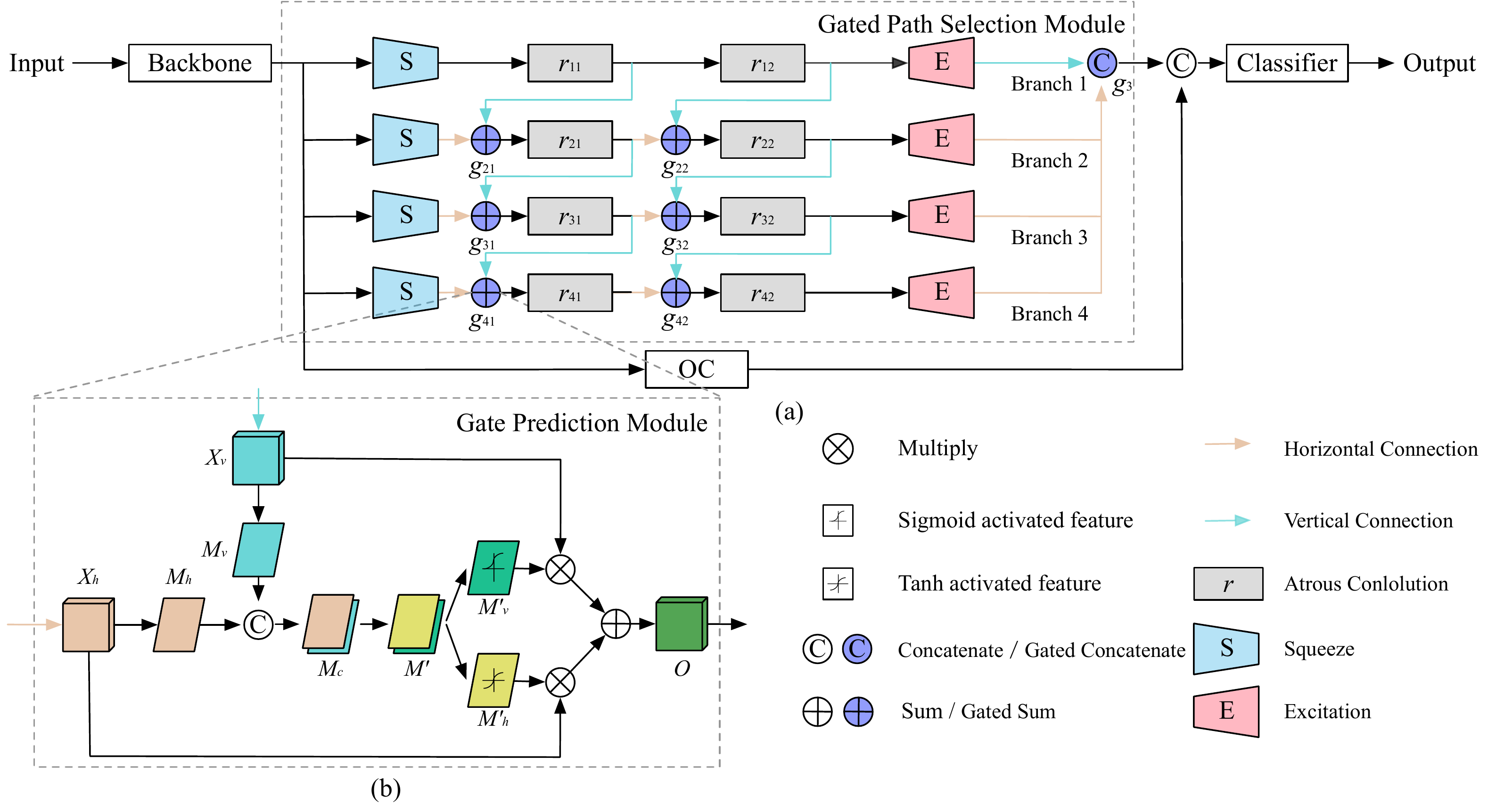}
    \caption{(a) \textbf{Overview of GPSNet}. Given an input image, a $1 \times 1$ convolution layer is used to squeeze the feature maps. Then we use atrous convlution layers with different dilation rates $r_{ij}$ to learn adaptive receptive field to sample features. The final features is excited by a $1 \times 1$ convolution layer. To get global context-aware features, we further integrate Object Context~(OC) module into our network. (b) \textbf{An illustration of gate prediction module}. It takes the vertical features $X_v$ and horizontal features $X_h$ as inputs, and estimates the soft masks \{$M'_v, M'_h$\}to reweight the original features by element-wise multiplication. Final features is produced through summing the reweighted features.}

    \label{fig:overview}
\end{figure*}

In this section, we present \namefull{}~(\name{}) for semantic segmentation in detail. \name{} estimates gate masks to weight contexts which are derived from various receptive fields~(RFs) and sample rates~(SRs).  Specifically, in \name{} we first carefully design a SuperNet with atrous convolution layers where the dilation rates varying in large range. Then a gate prediction module is introduced to determine the importance of samples from different feature maps and dynamically aggregate them.

\subsection{SuperNet} 
\paragraph{ASPP.} Atrous convolution is introduced to replace compact convolution in dense prediction. It is used to increase receptive field while maintaining the resolution of feature maps. However, atrous convolution fails to capture multi-scale semantic context. ASPP is proposed to concatenate feature maps in multiple parallel atrous convolution layers with different dilation rates. One advantage of ASPP is to sample features at different scales to boost the performance. However, as the dilation rate increases, the atrous convolution layer gradually loses the power of modeling and fails to capture information effectively~\cite{chen2017rethinking}. 

\paragraph{SuperNet.} To further enhance the capability to learn effective feature representations, we propose an improved ASPP-like network structure with multiple entrances and exits to propagate information among atrous convolution layers. We modify the original ASPP structure in the following three aspects.

\textit{Tuned Dilation.} We first extend the ASPP atrous convolution layers to grid form. We double the atrous convolution layers with dilation rates \{$r_1$, $r_2$, $r_3$, $r_4$\} into \{($r_{1}$, $r_{1}$), ($r_{2}$, $r_{2}$), ($r_{3}$, $r_{3}$), ($r_{4}$, $r_{4}$)\}. To mitigate repeatedly sampling, the dilation rates are further tuned to improve the sample rate. As illustrated in Figure~\ref{fig:overview}(a), we replace the ASPP dilation parameters with prime numbers which are \{1, 3, 11, 13, 23, 29, 33, 37\} to produce \{($r_{11}$, $r_{12}$), ($r_{21}$, $r_{22}$), ($r_{31}$, $r_{32}$), ($r_{41}$, $r_{42}$)\}. 

\textit{Bottlenecked Branch.} To alleviate the GPU memory usage and computing resource overhead, inspired by eASPP~\cite{valada2018self}, we introduce bottlenecked branches in SuperNet to address the problem. Following \cite{hu2018squeeze}, each of the branch starts with a \emph{Squeeze} operation which is to reduce the channel of input features with a $1\times1$ convolution. Then two consecutive $3\times3$ atrous convolutions are used to extract features with different sample rates. Finally, an \emph{Excitation} operation is applied at the exit of each branch, the features are produced to large channel features with a $1 \times 1$ convolution. All the convolutions are followed by InplaceABNsync~\cite{rota2018place}.

\textit{Dense Connectivity.} To facilitate information flow between different atrous convolution layers, we use dense connectivity to bridge parallel bottlenecked branches in SuperNet. The intermediate feature maps are aggregated by the results of one or two directions: 1) the output of a convolution from previous layer in the same branch~(horizontal connection) and, if possible, 2) the result of a convolution from the previous branch~(vertical connection).
Consequently, the subsequent layers gather the information from early layers with relative small receptive filed layers. In comparison with ASPP and DenseASPP, because of the dense connectivity pattern, we can acquire abundant features with more diverse and denser context.

\subsection{Gate Prediction Module}
In this subsection, we describe the gate prediction module in detail. The module is used to dynamically aggregate information in the SuperNet. In existing work, different features are merged through directly concatenation and summation. Actually, objects are in complex geometric transformations, augment features via simple concatenation or summation may be infeasible in dense prediction task.
Therefore, we propose a gate prediction module to composite different scales of features.
Figure~\ref{fig:overview} (b) depicts the process of the gate prediction module. It contains three operators: Projection, Comparison and Weighted Sum.


\textit{Projection.} Each gate considers two input feature maps $X_v, X_h$ with the shape of $H \times W \times C$, where $X_v$ is feature maps from the previous vertical layer and $X_h$ is the horizontal input features. The soft gate masks $M_v, M_h$ with the size of $H \times W \times 1$ are predicted via a projection transformation $\mathcal{F}$, where 

\begin{equation}
    \mathcal{F}: X_i \rightarrow M_i, i \in \{v, h\}.
\end{equation}

The transformation layer is defined with three consecutive operations: a $1 \times 1$ convolution, followed by a batch normalization~(BN) and a rectified linear unit~(ReLU). 

\textit{Comparison.} To adaptively adjust the pixel-wise receptive field and sample rates, the soft mask is used to integrate information from different scales of features. We first get the concatenation mask $M_c = [M_v, M_h]$ and a comparison function $\mathcal{C}_a$ is applied to get the soft mask $M' \in \mathbb{R}^{H \times W \times 2}$: 

\begin{equation}
    \mathcal{C}_a: M_c \rightarrow M'.
\end{equation}

The comparison function $\mathcal{C}_a$ in our experiments consists of a $1 \times 1$ convolution layer, followed by a BN layer and a ReLU layer. 

\textit{Weighted Sum.} To aggregate features, we first split $M'$ into $\{M_v', M_h'\}$ along the channel dimension, and reweight the input feature $\{X_v, X_h\}$ by element-wise multiplication. The adaptive features $O \in \mathbb{R}^{H \times W \times C}$ are obtained by summing the reweighted features.   

\begin{equation}
    O = M_v' \otimes X_v + M_h' \otimes X_h.
\end{equation}



\subsection{Relation to other approaches}
In this section, we compare our \name{} with the most relevant approaches including ASPP, Dense Atrous Spatial Pyramid Pooling~(DenseASPP), Deformable Convolution Network~(DCN). 

\begin{figure}[ht]
    \centering
    \includegraphics[width=\linewidth]{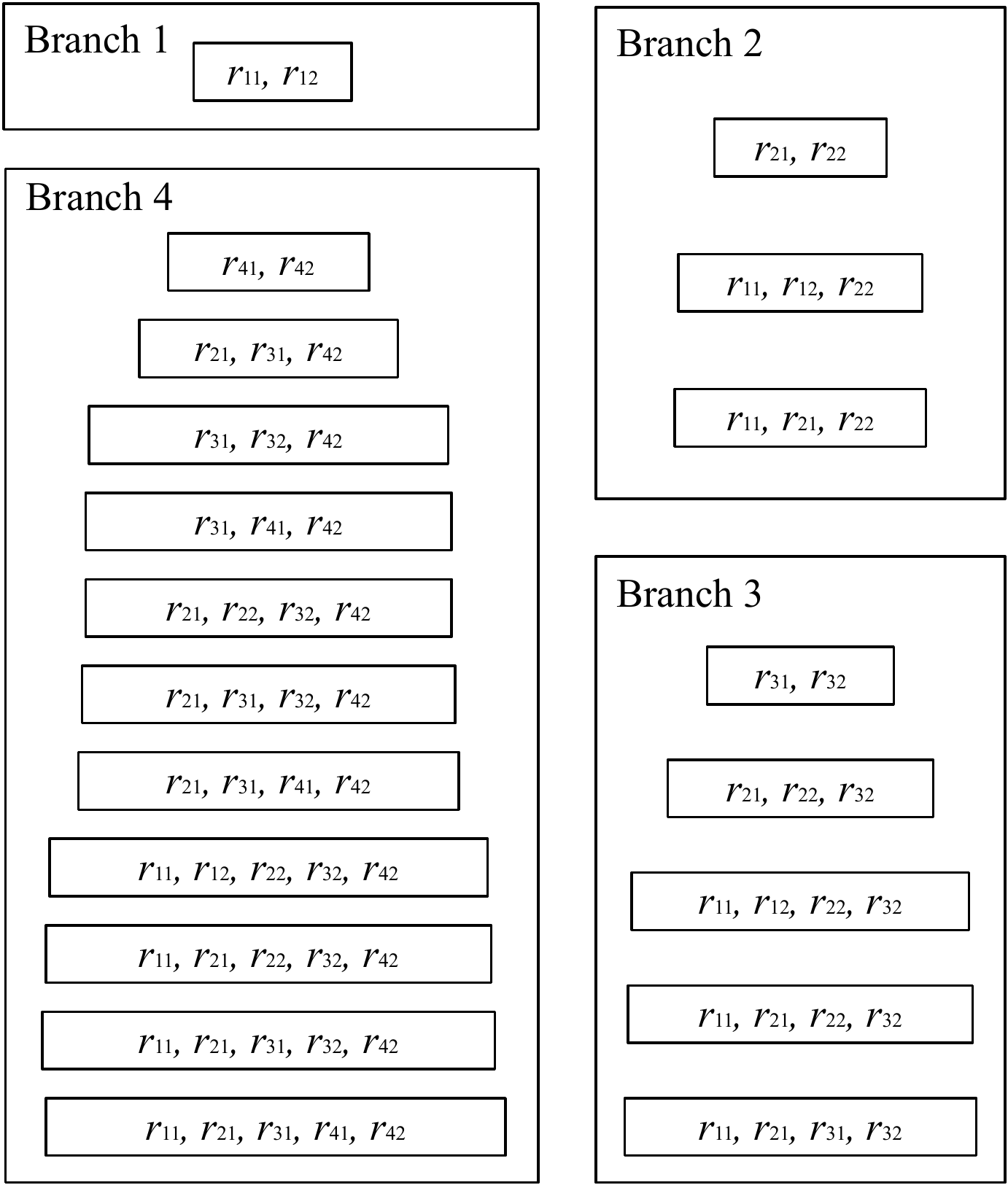}
    \caption{Illustration of the scale pyramid of each branch in SuperNet. Each branch produces feature pyramid with much larger receptive field and denser sample rate than ASPP.}\label{fig:unfold}
\end{figure}

\paragraph{ASPP.} ASPP~\cite{chen2017deeplab} adopts atrous convolution layers to segment both small and large objects. It employs multiple parallel filters with different rates to exploit multi-scale features. The features extracted from each sample rate are further concatenated to produce the final result. 
It is different from \name{} which applies soft gates to dynamically reweight the feature maps from separate branches. Moreover, by extending the parallel atrous convolution layers to grid form, \name{} expands the receptive fields to large variations. As illustrated in Figure~\ref{fig:unfold}, the sample positions in \name{} are linearly increased when more branches are added.

\paragraph{DenseASPP.} In order to achieve large enough receptive field size, DenseASPP introduces a base cascade network which consists of atrous convolution layers. The final results are obtained for an input visiting from small receptive field to large receptive field sequentially. Differently, \name{} introduced SuperNet with multiple entrances and multiple exits, which is more flexible to get different scales of features. Specifically, we can feed the input in any entrances and reject it from different exits. Through gated modules, our method can generate adaptive receptive fields to tackle objects with large geometric deformations.

\paragraph{DCN.} To make the convolution kernel adaptive to geometric variations of the object, DCN predicts 2D offsets to augment spatial sampling positions. In contrast, \name{} predicts soft gates to select the semantic context for objects with different shapes. It can get large receptive field, and enable densely position samplings in the effective semantic context.


\section{Experimental Results}
To demonstrate the effectiveness of our approach, we conducted extensive experiments on two representative semantic segmentation datasets, \ie, Cityscapes~\cite{cityscapes}, ADE20K~\cite{ade20k}.
In addition, complete ablation studies are performed to analyze the components of \name{}. To integrate global-aware information, we incorporate object context~(OC) module~\cite{ocnet} in our network. All of our experiments are conducted in PyTorch. The code to reproduce our results will be made public upon the acceptance of the paper. 

\subsection{Evaluation on Cityscapes}

\paragraph{Dataset.} Cityscapes is the dataset to understand urban scenes. It contains 30 common classes including road, person, car, \etc. and only 19 of them are used for semantic segmentation evaluation. The dataset is comprised of 5,000 finely annotated images and 20,000 coarsely annotated images. The finely annotated 5,000 images are divided into 2,975, 500 and 1,525 images for training, validation, and testing. 

\paragraph{Training Details.}
On the Cityscapes dataset, we train all models with the 2,975 finely annotated images. We set the mini-batch size as 8 with InplaceABNSync~\cite{rota2018place} to synchronize the mean and standard variation. The initial learning rate is set as 0.01 and weight decay as 0.0005. Following PSPNet~\cite{pspnet}, the original image is randomly cropped to produce $769 \times 769$ input. We train models with 40K iterations with 4$\times$P40 GPUs. Following previous work~\cite{pspnet}, we employ the `poly' learning rate policy, where the power is set to 0.9. We augment the dataset by scaling it with a factor in the rage of [0.5, 2], horizontally flipping.

\paragraph{Ablation Study.}
\begin{table*}[h]
    \centering
    \begin{tabular}{c | c c c c | c}
    \hline
        ~&~SuperNet~&~Gate Prediction Module~&~Tuned Dilation~&~OHEM~&~Mean IoU~($\%$) \\
    \hline
    \multirow{2}{*}{ASPP}~&~-~&~-~&~-~&~-~&~78.65 \\
    &~\checkmark~&~\checkmark~&~\checkmark~&~-~&~79.31 \\
    \hline
    \multirow{5}{*}{OCNet}~&~-~&~-~&~-~&~-~&~79.58~(78.70) \\
    &~\checkmark~&~-~&~-~&~-~&~79.71 \\
    &~\checkmark~&~\checkmark~&~-~&~-~&~80.03 \\
    &~\checkmark~&~\checkmark~&~\checkmark~&~-~&~80.32 \\
    &~\checkmark~&~\checkmark~&~\checkmark~&~\checkmark~&~81.21 \\
    \hline
    \end{tabular}
    \caption{Ablation study on Cityscapes validation dataset. The improvements of individual components of the proposed approach are evaluated. We report the results of OCNet from literature~(79.58) and our reproduced experiments~(78.70).}
    \label{tab:ablation}
\end{table*}


To investigate the effectiveness of the individual components of the proposed approach, \ie, SuperNet, Gate Module, Tuned Dilation and online hard example mining(OHEM)~\cite{wu2016high}. We integrate the components into two representative approaches, \ie, ASPP, OCNet. The ablation analysis is conducted on the Cityscapes validation set. Quantitative results are shown in Table~\ref{tab:ablation}. In Table~\ref{tab:ablation}, we show the mIoU 79.58 reported by OCNet, which is reproduced to be 78.70 in our experiment. All the components consistently improve the performance of both approaches.

\begin{itemize}
\item{\textbf{SuperNet.}} To further evaluate the effectiveness of SuperNet, we compare the different OCNet trained with SuperNet and with the baseline OCNet. Validation accuracy in both settings is shown in Table~\ref{tab:ablation}. With SuperNet, the validation accuracy is higher than the baseline model. This demonstrates SuperNet with different scales of receptive fields can help improve performance.

\item{\textbf{Gate Prediction Module}} is used to adaptively select receptive fields. It can improve the performance as shown in Table~\ref{tab:ablation} trained with OCNet. The SuperNet typically benefit from the gate module. The larger receptive field layers acquire information from the previous layers with relative small receptive fields. The gate module is able to not only control the size of the field of view but also increase the sample rate within the effective receptive field. This indicates that adaptive receptive fields are of importance in dense object prediction to deal with object transformations.

\item{\textbf{Tuned Dilation.}} Quantitative improvements with the tuned dilation as shown in Table~\ref{tab:ablation}. By integrating the carefully tuned atrous convolution layers, the network tends to a gain higher sample rate which enables to densely capture the semantic context. The result further shows that tuned dilation is indeed effective for increasing sample rate to improve the performance.

\item{\textbf{OHEM.}} To tackle with data imbalance and overfitting, we further conduct experiments to compare the models with OHEM and baseline in Table~\ref{tab:ablation}. following previous work~\cite{wu2016high}, we set the threshold for selecting hard pixels as 0.7, and keep at least 10,0000 pixels within each mini-batch. The result shows that the OHEM built on our network can further boost the performance.
\end{itemize}

\paragraph{Performance.} On Cityscapes, we compare \name{} with several competitive baselines including the dilation-based methods, \ie, DeepLabv3~\cite{chen2017rethinking}, DUC-HDC~\cite{wang2018understanding}, DenseASPP~\cite{denseaspp}, region-based method \ie, PSPNet~\cite{pspnet}, and attention-based method \ie, PSANet~\cite{psanet}, OCNet~\cite{ocnet}. We evaluate our results on the Cityscapes testing set with multi-scale testing. Results are shown in Table~\ref{tab:cityscapes}. The prediction of \name{} is substantially more accurate than the methods conducted with ResNet-101. In addition, our result also outperforms DenseASPP which takes DenseNet-161 as backbone. Visual results are shown in Figure~\ref{fig:vis}.


\begin{table}[ht]
    \centering \scalebox{1.0} {
    \begin{tabular}{c|c|c}
    \hline
    Method & BaseNet & Mean IoU~($\%$) \\
    \hline \hline
    DenseASPP~\cite{denseaspp} & DenseNet161 & 80.6 \\
    \hline
    Deeplabv3$^*$~\cite{chen2018encoder} & ResNet101 & 81.3 \\ 
    DUC-HDC~\cite{wang2018understanding} & ResNet101 & 77.6 \\
    PSPNet~\cite{pspnet} & ResNet101 & 78.4 \\
    PSANet~\cite{psanet} & ResNet101 & 78.6 \\
    OCNet$^\dagger$~\cite{ocnet} & ResNet101 & 80.1 \\ 
    OCNet & ResNet101 & 81.2 \\ 
    \hline
    \name{}$^\dagger$ & ResNet101 & 80.6 \\
    \name{} & ResNet101 & \textbf{82.1} \\
    \hline
    \end{tabular}}
    \caption{Results on Cityscapes test dataset. The results marked with $\dagger$ indicate the models trained without validation set. The result of Deeplabv3 marked with $*$ is trained with both finely and coarsely annotated training data.}
    \label{tab:cityscapes}
\end{table}

\subsection{Evaluation on ADE20K}
\paragraph{Dataset.}The scene parsing dataset ADE20K contains 150 classes and diverse complex scenes with 1,038 image-level categories. It needs to parse both objects and stuff. The dataset is divided into 20,000, 2,000 and 3,000 for training, validation and testing. Results are evaluated with both \emph{mean of class-wise Intersection over Union}~(Mean IoU).

\paragraph{Training Details.} On the ADE20K dataset, the base learning rate is set as 0.02 and with a weight decay 0.0001. The input image is resize to $480 \time 480$. The mini-batch size is 16 and we also apply InplaceABNSync to synchronize the mean and standard deviation across multiple GPUs. The models are trained with 200K iterations with 4$\times$P40 GPUs. The learning rate policy and data augmentation are the same as the Cityscapes dataset. 

\begin{figure}[ht]
    \centering
    \includegraphics[width=\linewidth]{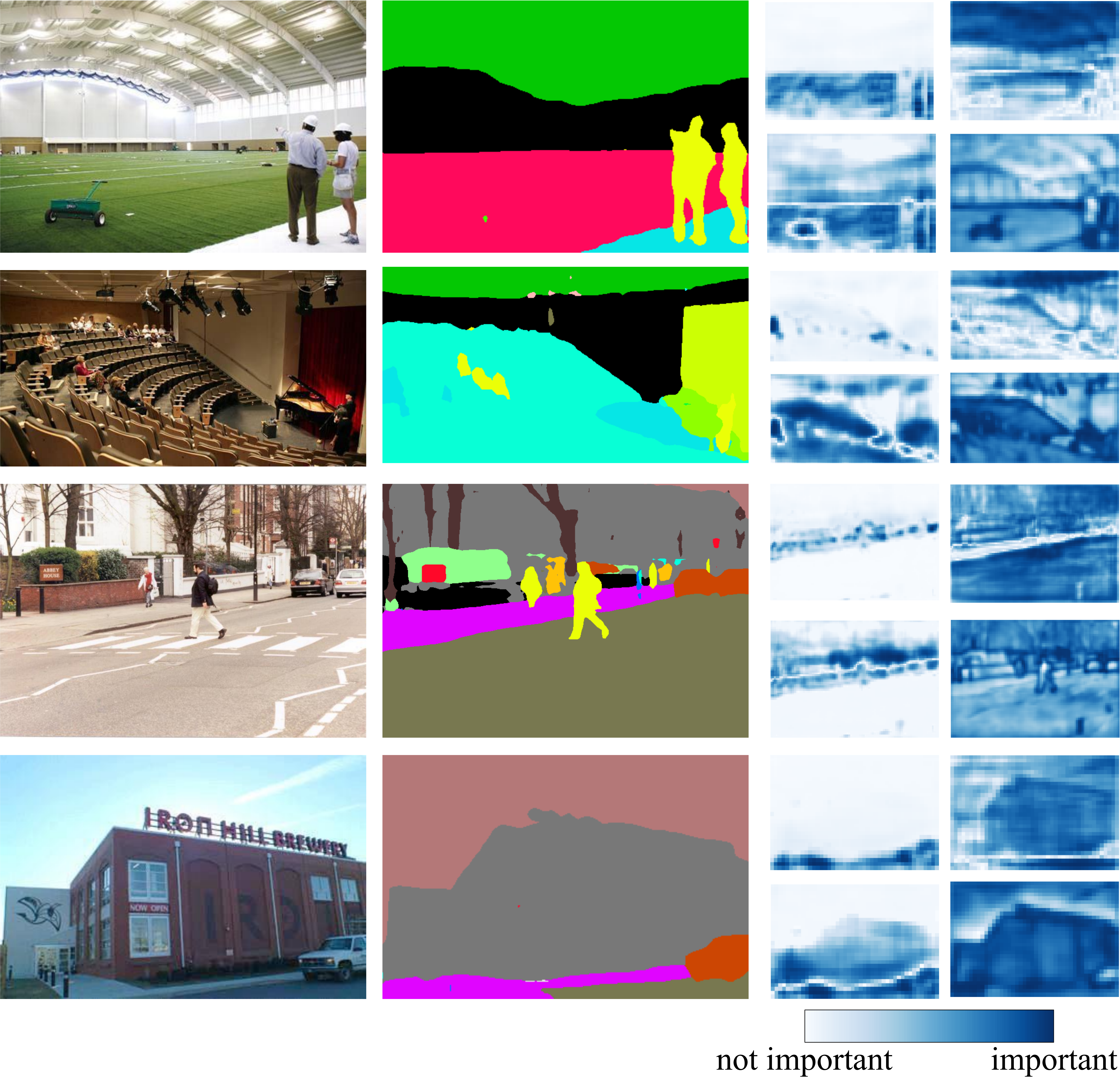}
    \caption{Visualization of the normalized estimated soft masks. The third column is the masks from $g_{31}$. The fourth column is the masks from branch 3 and branch 4.}\label{fig:vis_ade}
\end{figure}

\begin{figure*}[ht]
    \centering
    \includegraphics[width=\linewidth]{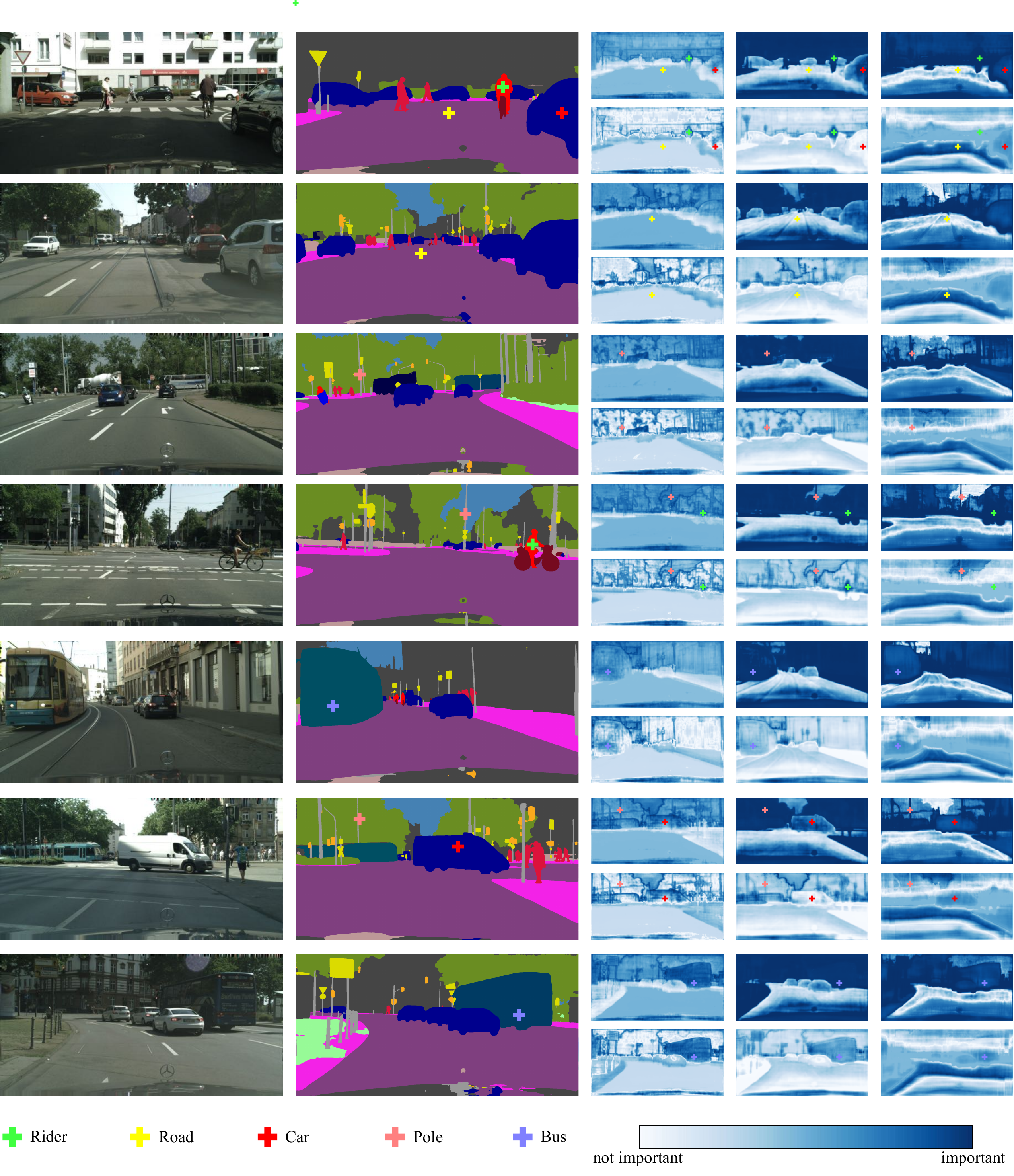}
    \caption{Visualization of the normalized estimated soft masks. The third column is the masks from $g_{21}$. The fourth column is the masks from $g_{22}$, and the last column is the masks from branch 1 and branch 4.}\label{fig:vis}
\end{figure*}

\paragraph{Performance.} On ADE20K, we compare our evaluated \name{} with three attention-based method, \ie, PSANet~\cite{psanet}, EncNet~\cite{encnet}, OCNet~\cite{ocnet}, and region-based method, \ie, PSPNet. The experiments are evaluated on the ADE20K validation set. The results reported in Table~\ref{tab:ade20k} shows that \name{} consistently outperform all baselines. Notable, \name{} surpasses a 269-layer PSPNet. Visual results are shown in Figure~\ref{fig:vis_ade}.

\begin{table}[ht]
    \centering
    \begin{tabular}{p{2cm}<{\centering}|p{1.5cm}<{\centering}|p{2.3cm}<{\centering}}
    \hline
    Method & BaseNet & Mean IoU~($\%$) \\
    \hline \hline
    PSPNet~\cite{pspnet} & ResNet269 & 44.94 \\
    \hline
    PSPNet & ResNet101 & 43.29 \\
    PSANet~\cite{psanet} & ResNet101 & 43.77 \\
    EncNet~\cite{encnet} & ResNet101 & 44.65 \\
    OCNet~\cite{ocnet} & ResNet101 & 45.45 \\
    \hline
    \name{} & ResNet101 & \textbf{45.76} \\
    \hline
    \end{tabular}
    \caption{Results on ADE20K validation dataset.}
    \label{tab:ade20k}
\end{table}

\subsection{Model Analysis}
\begin{table*}[h]
    \centering
    \begin{tabular}{c|c|c|c|c}
    \hline
    Method & Dilation Setting & RF & SR & \#Params. \\
    \hline
    ASPP & \{1,12,24,36\} & 73 & 0.006 & 18.9M \\ 
    DenseASPP & \{1,12,24,36\} & 147 & 0.070 & 25.6M \\
    SuperNet & \{(1, 1),(12, 12),(24, 24), (36, 36)\} & 219 & 0.125 & 6.29M \\
    Untuned GPS & \{(1, 1),(12, 12),(24, 24), (36, 36)\} & 219 & 0.125 & 6.3M \\
    Tuned GPS & \{(1, 3),(11, 13),(23, 29), (33, 37)\} & 199 & 0.843 & 6.3M \\
    \hline
    \end{tabular}
    \caption{Comparing GPS module with other methods with the same dilation rates setting, GSP module provides larger RF, higher SR and introduces less parameters.}
    \label{tab:complex}
\end{table*}

In this section, we analyze our approach in three aspects, \ie, receptive field, sample rate and parameter efficiency. The statistics of the size of receptive field, sample rate and the number of parameters of different methods are summarized in Table~\ref{tab:complex}.

\paragraph{Receptive Field~(RF).}
The result in the table shows that with the same dilation settings, the size of receptive fields of our method is substantially improved $99\%$ and $49\%$ over ASPP and DenseASPP respectively.

\paragraph{Sample Rate~(SR).}
\begin{figure}[t]
    \centering
    \includegraphics[width=\linewidth]{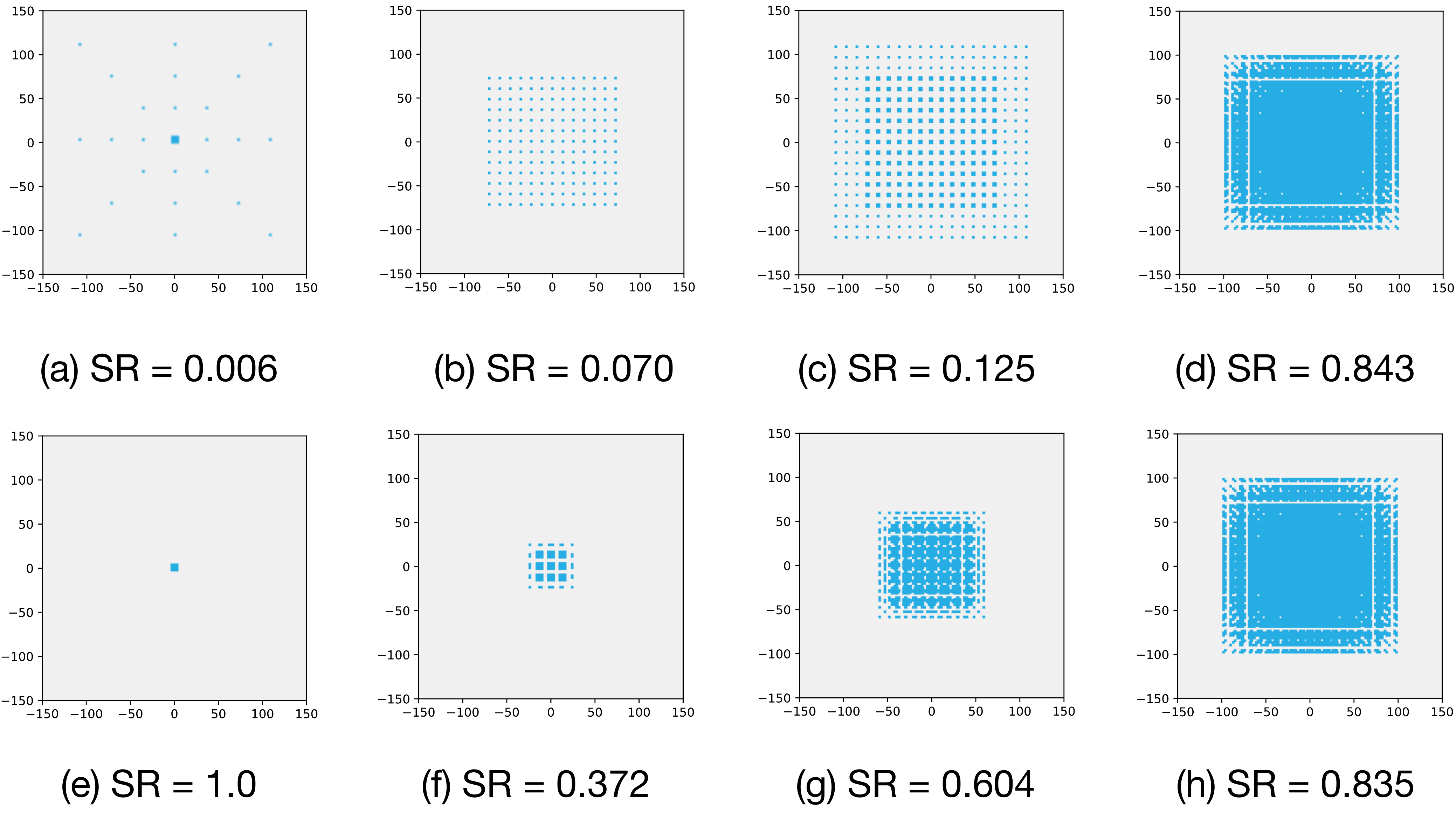}
    \caption{Illustration of sample position of (a) ASPP, (b) DenseASPP, (c) Untuned SuperNet, (d) Tuned SuperNet, (e) Branch 1 of (d), (f) Branch 2 of (d), (g) Branch 3 of (d) and (h) Branch 4 of (d). 
    }\label{fig:rf}
\end{figure}

In Figure~\ref{fig:rf}, we compare the sample rate of different methods. The results presented in the figure demonstrate that SuperNet in \name{} provides much higher SR. With the untuned dilation setting, our SR is 0.125 which is almost twice higher than DenseASPP. After tuning the ASPP grid parameters, SR is further increased to 0.843.

\paragraph{Parameter Efficiency.}
The results in Table~\ref{tab:complex} indicate that SuperNet utilizes parameters more effectively than alternative methods, \ie, ASPP and DenseASPP. This suggests that \name{} is simple but effective to extract local context handle object with geometric variations 

\subsection{Adaptive Receptive Fields Visualization}

As shown in Figure~\ref{fig:vis}, we choose pixels from rider, road, car, pole and bus to visualize the path selection mechanism. The masks are relative with category, scales and appearance variance frequency. There are several conclusions as follows.
\begin{itemize}
\item For movable objects such as car and rider, the main features are captured with branch 1, in which the RF is the smallest.
\item For textureless regions such as road, the features are captured with branch 4, in which the RF is the largest. 
\item To aggregate boundary features, the features from the convolution layer with dilation rate $r_{11}$ are estimated with large weights.
\item For large objects like bus, the features from branch 4 are given larger weight than small objects.
\item For tiny objects like pole, the local features from branch 1 are much more important than the features from branch 4.
\end{itemize}

\section{Conclusion}
In this paper, we have presented \emph{\namefull{}~(\name{})} to learn adaptive receptive fields and increase sample rates in semantic segmentation. 
A SuperNet is proposed to ensure maximum information flow in the network. It provides various paths to extract multi-scale representations. Dense connectivity in the network allows to feed the input from small receptive field to large receptive field.
A gate prediction module is further introduced to estimate soft masks to dynamically select effective context positions and regularize the receptive fields. 
Besides, our method is simple, efficient and model-agnostic. It can be applied to various ASPP-like architectures. 
The proposed method has shown its effectiveness on two competitive semantic segmentation datasets, \ie, Cityscapes, ADE20K, and achieves new state-of-the-art results.
Future research may focus on extending our results
to other types of computer vision tasks, such as object detection and image generation.

{\small
\bibliographystyle{ieee_fullname}
\bibliography{citation}
}
\clearpage
\appendix 
\section{Receptive Field~(RF) and Sample Rate~(SR)}
In this section, we quantify receptive field and sample rate of different approaches in detail.


\paragraph{Atrous Convolution.} The receptive field and sample rate of atrous convolution are defined in Eqn.~\ref{eq:1}, where $r$ and $k$ are the dilation rate and kernel size respectively. 

\begin{equation}
\begin{aligned}
    \text{RF}_\text{ac} &= (r \times k - r + 1)^2, \\
    \text{SR}_\text{ac} &= \frac{k^2} {\text{RF}_\text{ac}}.
\end{aligned}
\label{eq:1}
\end{equation}


\paragraph{ASPP} adopts atrous convolution layers with different dilation rates to extract multi-scale context. The receptive field and sample rate are defined in Eqn.~\ref{eq:2}, where $B$ is the number of branches, and $b$ indicates the index of branch. 
There is a clear trend that the number of sample positions is linearly correlated with the number of branches.

\begin{equation}
\begin{aligned}
    \text{RF}_\text{aspp} &= (max(r_b)\times (k-1)+1)^2, \\
    \text{SR}_\text{aspp} &= \frac{(B \times k^2-B+1)} {\text{RF}_\text{aspp}}.
\end{aligned}
\label{eq:2}
\end{equation}

\paragraph{DCN} introduces convolution layers to predict offsets to dynamically adjust the receptive fields. Eqn.~\ref{eq:3} gives the definition of receptive field and sample rate:

\begin{equation}
\begin{aligned}
    \text{RF}_\text{dcn} &= (max(p_{i,x})-min(p_{i,x}) \\ & \times (max(p_{i,y})-min(p_{i,y}), \\
    \text{SR}_\text{dcn} &= \frac{k^2} {\text{RF}_\text{dcn}},
\end{aligned}
\label{eq:3}
\end{equation}

Where $(p_{i,x}, p_{i,y})$ is the position of the $i^{th}$ sample.

\paragraph{A Serial of Atrous Convolution.} We define a serial of atrous convolution: $S = ((k_1, r_1),(k_2, r_2),...,(k_n, r_n))$. The sample position set $\{s_i\}$ can be obtained by walking through layers. The receptive field and sample rate are formulated in Eqn.~\ref{eq:serial}.

\begin{equation}
\begin{aligned}
    \text{RF}_\text{s} &= (\sum_i(r_i \times (k_i-1)) + 1)^2, \\
    \text{SR}_\text{s} &= \frac{| \{s_i\} |} {\text{RF}_\text{s}}.
\end{aligned}
\label{eq:serial}
\end{equation}


\paragraph{DenseASPP} densely connects a set of atrous convolution layers. Given dilation rates $\{r_1, r_2, r_3, r_4\}$, for level $l$ in the DenseASPP pyramid, the sample set is $P_l = \{s_i\}_{l}$, which is the union of sample positions. The receptive field and sample rate are: 

\begin{equation}
    \begin{aligned}
        \text{RF}_\text{denseaspp} &= (\sum_i(r_i \times (k_i-1)) + 1)^2, \\
        \text{SR}_\text{denseaspp} &= \frac{| \bigcup_l (\{P_l\}) |} {\text{RF}_\text{denseaspp}}.
    \end{aligned}
    \label{eq:serial}
\end{equation}

\paragraph{GPSNet.} In Figure 3~(\textbf{in paper}), we unfold all the GPS branches. For a specific output position, four pyramids with different scales are used to extract features, and each of the pyramid corresponds to a branch. Hence, the sample position set of each branch is $P_b = \bigcup_l (\{P_{b,l}\}))$. The RF and SR of GPS module can be calculated as follows:

\begin{equation}
    \begin{aligned}
        \text{RF}_\text{gps} &= max(RF_{b\in\{1,2,3,4\}}), \\
        \text{SR}_\text{gps} &= \frac{| \bigcup_b(\{P_b\})|} {\text{RF}_\text{gps}}.
    \end{aligned}
    \label{eq:4}
\end{equation}


\begin{figure*}[t]
    \centering
    \includegraphics[width=\linewidth]{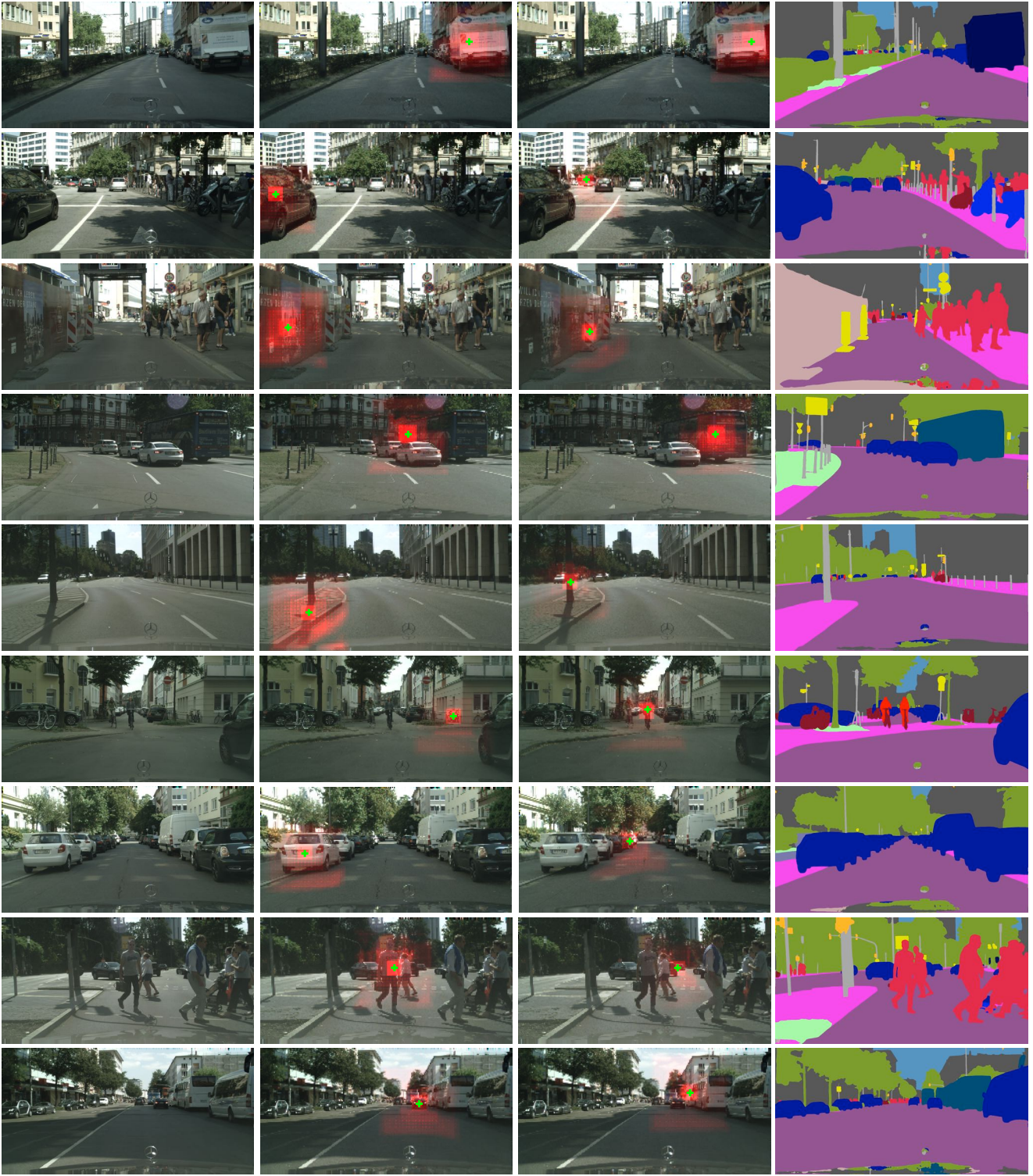}
    \caption{Visualization of our results. In column 2 and column 3, the selected positions are marked with green dot, and the regions with dark red are more relevant to the corresponding positions.}
    \label{fig:sup_vis}
\end{figure*}

\vspace{-0.5cm}

\section{Understanding GPSNet}
\name{} is built on the idea of weighting different sample locations through soft gates to get free form receptive fields.
Comparing to the fixed RFs and SR over the feature map, the position-wise weights are gradually used from small RF to large RF to gather related context. 

Instead of estimating irrelevant sample positions, the inner and surrounding semantic context are selected effectively as shown in Figure~\ref{fig:sup_vis}. The main contribution to extract semantic context is mainly from the positions within the objects, which is consistent with the observation in \cite{luo2016understanding}. Meanwhile, the surrounding compatibility contexts like road or rider, are also served as complement to eliminate confusing ambiguity.
\end{document}